\documentclass{osa-article}

\journal{osac}

\articletype{Research Article}

\usepackage{lineno} 

\usepackage{adjustbox} 

\usepackage{xcolor,colortbl} 
\arrayrulecolor{black}

\usepackage{soul} 

\newcommand{\revise}[1]{\textcolor{blue}{{#1}}}

\definecolor{blue}{rgb}{0,0,0} 
\newcommand{\delate}[1]{\textcolor{blue}{}}
\newcommand{\delrev}[1]{\textcolor{blue}{}}

\begin{document}

\title{Turbulence Strength $C_n^2$ Estimation from Video using Physics-based Deep Learning}

\author{Ripon Kumar Saha,\authormark{1,*} Esen Salcin,\authormark{2} Jihoo Kim,\authormark{2} Joseph Smith,\authormark{2} and Suren Jayasuriya,\authormark{1}}

\address{\authormark{1} Arizona State University, Tempe, AZ 85281, USA\\
\authormark{2} Alphacore Inc, 304 S Rockford Dr, Tempe, AZ 85281, USA}

\email{\authormark{*}rsaha8@asu.edu} 



\begin{abstract}


Images captured from a long distance suffer from dynamic image distortion due to turbulent flow of air cells with random temperatures, and thus refractive indices. This phenomenon, known as image dancing, is commonly characterized by its refractive-index structure constant $C_n^2$ as a measure of the turbulence strength. For many applications such as atmospheric forecast model, long-range/astronomy imaging, and aviation safety, optical communication technology, $C_n^2$ estimation is critical for accurately sensing the turbulent environment. Previous methods for $C_n^2$ estimation include estimation from meteorological data (temperature, relative humidity, wind shear, etc.) for single-point measurements, two-ended pathlength measurements from optical scintillometer for path-averaged $C_n^2$, and more recently estimating $C_n^2$ from passive video cameras for low cost and hardware complexity. In this paper, we present a comparative analysis of classical image gradient methods for $C_n^2$ estimation and modern deep learning-based methods leveraging convolutional neural networks. To enable this, we collect a dataset of video capture along with reference scintillometer measurements for ground truth, and we release this unique dataset to the scientific community. We observe that deep learning methods can achieve higher accuracy when trained on similar data, but suffer from generalization errors to other, unseen imagery as compared to classical methods. To overcome this trade-off, we present a novel physics-based network architecture that combines learned convolutional layers with a differentiable image gradient method that maintains high accuracy while being generalizable across image datasets. 
\end{abstract}

\section{Introduction}
In remote sensing or long-distance imaging, atmospheric turbulence can significantly degrade image quality with optical distortion and blur. This effect is caused by changes in the refractive index of the medium inducing bending of light rays in propagation. Turbulence and its strength vary both spatially and temporally, and is represented by the refractive index structure constant $C_n^2$ (typically measured as path-averaged with optical equipment). $C_n^2$ has units of $m^{-2/3}$ and can vary from $10^{-16}$ to $10^{-12}$ for a typical day depending on the environmental conditions. Accurate estimation of turbulence strength is useful when assessing the performance of the optical system in a real-world scenario and to alleviate turbulence effects for images captured in operation.

Path averaged $C_n^2$ is most accurately measured with scintillometers that leverage a double-ended optical setup consisting of a transmitter and a receiver at the opposite ends of the turbulent media to observe scintillation. However, such optical devices can be bulky and costly, and setting up the two-ended measurement may not be convenient for slant paths and moving source and/or target platforms. In contrast, there have been methods proposed to estimate $C_n^2$ passively from video cameras ~\cite{porat2011optical, o2017strength} which could have the potential to be used in a single-ended manner without requiring a known target for imaging. These algorithms calculate the gradient of an image and derive physics-based equations (under simplifying assumptions) for the turbulence strength. However, it is unclear how accurate these methods can achieve, and what factors of variation in the optical images can affect their performance. For instance, Porat et al. estimated linearity (consistency of measurements over the entire range of measurements) of 0.8 to 0.58 based on med-weak to strong turbulence and estimated standard deviation of the error (variability across all samples) ranging from 0.59 to 0.96 between the estimated value and scintillometer values~\cite{porat2010crosswind}. Zamek et al. estimated an error of $5\times10^{-15}$ to $1.5\times10^{-12}$ based on a single morning experiment~\cite{zamek2006turbulence, zamek2006TurSuper}. \revise{Note that all these papers utilize different metrics and did not provide their raw data to compute a common metric, making it difficult to compare across the papers. However, different metrics correspond to different performance criteria: linearity/correlation coefficient corresponds to how well the turbulence strength trend is predicted over time, standard deviation represents the variability of the measurement, and mean absolute error represents the accuracy.}

There has been interest in applying modern deep learning techniques for their potential for achieving higher accuracy than the prior literature, although this research approach is currently bottlenecked by the lack of available datasets and ground truth $C_n^2$ measurements required for supervised learning. In this paper, we present a framework as we show in code 1 (Ref. \cite{saha}) for comparative study of both classical and deep learning algorithms for $C_n^2$ estimation. In particular, we collect an open-source dataset 1 (Ref. \cite{saha_dataset}) of video, collected with a telephoto lens, co-located with a reference scintillometer measurement for several hours. Leveraging this dataset, we study the robustness of the method from gradient-based algorithms~\cite{zamek2006turbulence, zamek2006TurSuper, porat2011optical, porat2010crosswind} with respect to motion stabilization, gradient choice, and hyperparameters. In addition, we perform supervised learning utilizing a convolutional neural network to directly regress $C_n^2$ values from image frames. We analyze the performance of this network including its ability (or lack thereof) to generalize to new datasets without training. Based on our observations, we design a hybrid neural network combined with a differentiable image gradient method~\cite{zamek2006turbulence, zamek2006TurSuper, porat2011optical, porat2010crosswind} that achieves better performance than classical methods but is able to generalize across datasets. This is the first work to apply deep learning to $C_n^2$ estimation from passive RGB video, and we open-source our data and code for scientific reproducibility.

\section{Related Work}
\label{sec:related}

Turbulence estimation is a long-studied subject with decades of research. Most previous methods utilize custom optical devices such as lasers, reflective corner cubes, and scintillometers or meteorological sensors. We outline these methods as well as methods that estimate $C_n^2$ passively from optical images (which we focus on in this paper). Note that we do not discuss the related area of mitigating or removing turbulent effects in images (e.g. Mao et al.~\cite{mao2020image}, Li et al.~\cite{li2021unsupervised}) which is not in the scope of this paper.  

\paragraph{Active optical measurements.} One optical approach uses a laser beam projected through a corner cube (a special arrangement of 3 mirrors to redirect light back to the source), and measured the return power to estimate the path-averaged $C_n^2$~\cite{cole2010measured}. This method performed well in summer conditions, but deteriorated for winter measurements~\cite{cole2009path}. Recently, a convolutional neural network based architecture was developed to get the image of a laser beam projected back from a corner cube (7km away) and estimate the $C_n^2$, which shows a great correlation in the case of higher strength~\cite{vorontsov2020atmospheric}. However, when the turbulence intensity is lower, because of lower fluctuation/variance of air, the image lacks diversity which leads to more errors in estimated results.  In astronomical imaging, vertical turbulence estimation can be done by Rayleigh beacon involving a powerful laser~\cite{zuraski2021vertical} mostly used in observatories. However, this approach is not suitable for turbulence measurement of horizontal or slant path profiles on Earth. Finally, a Hartmann Turbulence Sensor (an optical system) leveraging two laser sources is also capable to estimate several atmospheric turbulence parameters, such as Greenwood frequency, Fried’s coherence diameter, and the inner scale of turbulence~\cite{bose2018profiling}.

\textit{Scintillometers.} To measure atmospheric turbulence accurately, scintillometers are the most popular devices. Scintillometers jointly measure the power-law exponent and the structure constant of atmospheric turbulence, solving simultaneous equations formed by the irradiance scintillation index and fluctuation variance of angle-of-arrival(AOA)~\cite{gao2018joint}. Scintillometer data collection requires precise alignment of transmitter and receiver, which is challenging to achieve over kilometers of distance. Within the same brand scintillometer, the average measurement difference can be $18$ to $21\%$ ~\cite{kleissl2008large, kleissl2009scintillometer} with proper optical alignment. This difference can be as large as $35\%$ to $240\%$ ~\cite{van2011analysis} between each large aperture scintillometer produced by same manufacturer (Kipp \& Zonen). It was suspected this error can be be caused by poor focal alignment of the transmitter and receiver detector that causes ineffective use of the Fresnel lens used in scintillometer. This measurement error can only be fixed by comparing with a reference scintillometer in the field ~\cite{van2011analysis}. 


\paragraph{$C_n^2$ from Meteorological Measurement.} Alternative approache to active optical measurement is to derive relationships between $C_n^2$ and meteorological data. Basu et al. discussed the relationship between surrounding temperature and  $C_n^2$ profiles in the lower atmosphere~\cite{basu2015simple}. Research has measured single point $C_n^2$ (instead of path averaged $C_n^2$) from surface fluxes derived from single-level routine weather data containing Temperature, humidity, and joint structure parameter~\cite{van2014estimation}. 

For path-averaged $C_n^2$ measurement, a method was proposed to estimate the refractive index structure parameter over land given two vertical levels of conventional wind speed, temperature, and humidity info as input to the model~\cite{tunick2003cn2}. An artificial neural network was also proposed to estimate $C_n^2$ from meteorological variables (temperature, relative humidity, pressure, potential temperature gradient, wind shear)~\cite{wang2016using}. A recent approach also considers meteorological variables (temperature, relative humidity, wind velocity) at different heights (from the surface) combined with pressure and surface temperature to forecast $C_n^2$ on Antarctica~\cite{yang2021estimation}. However, most of these methods leverage surrounding information of single or two points which are not enough for the estimation of true path-averaged $C_n^2$.


\paragraph{$C_n^2$ from passive video.} Of particular interest for this paper is methods that estimate $C_n^2$ passively from video or time-lapse imagery. Previous research uses phase-based techniques to compute path weighted $C_n^2$ and Fried's coherence diameter~\cite{bose2018estimation, mccrae2017estimation}. Video-recorded data was used to measure crosswind based on temporal correlations of the intensity fluctuations of a naturally illuminated scene induced by atmospheric turbulence~\cite{porat2010crosswind}. 

The method most relevant for our study takes atmospherically degraded image sequences as input and calculates turbulence-induced spatiotemporal movements across the frames of the image set in statically high variance areas~\cite{zamek2006turbulence, zamek2006TurSuper}. We label this method the image gradient or gradient method for short throughout the paper. Porat et al.~\cite{porat2011optical} also used a similar approach to capture image sequences by video camera and calculate the variance of the angle of arrival(AOA) which is directly correlated to path averaged $C_n^2$. During their field experiment, they collected the ground truth scintillometer data from a two-station scintillometer system with 60m~100m optical length~\cite{zamek2006turbulence, zamek2006TurSuper, porat2011optical} where the distance from the camera to the image plane was a few hundred to thousands of meters. Different positioning of scintillometer and imaging device can induce some error as turbulence is \revise{spatially} variant and changes based on surface geometry or building~\cite{tunick2005characterization}. However, deploying several heavy scintillometers at different places to collect different datasets is often unpractical. 


\section{Data Collection and Method}

In this section, we first describe our real data collection including optical setup and experiment parameters. We then proceed to discuss our method and implementations of the image gradient method~\cite{zamek2006turbulence}, our baseline convolutional neural network (CNN) for turbulence strength estimation, and our hybrid physics-based CNN. All codes that have been implemented are made publicly available as we show in Code 1 (Ref. \cite{saha}). Besides, the July and October dataset is available as we show in Dataset 1(Ref. \cite{saha_dataset}).



\subsection{Data Collection}

\begin{figure}[h!]
    \centering
    \begin{subfigure}{0.49\columnwidth}
    \centering\includegraphics[width=.99\columnwidth]{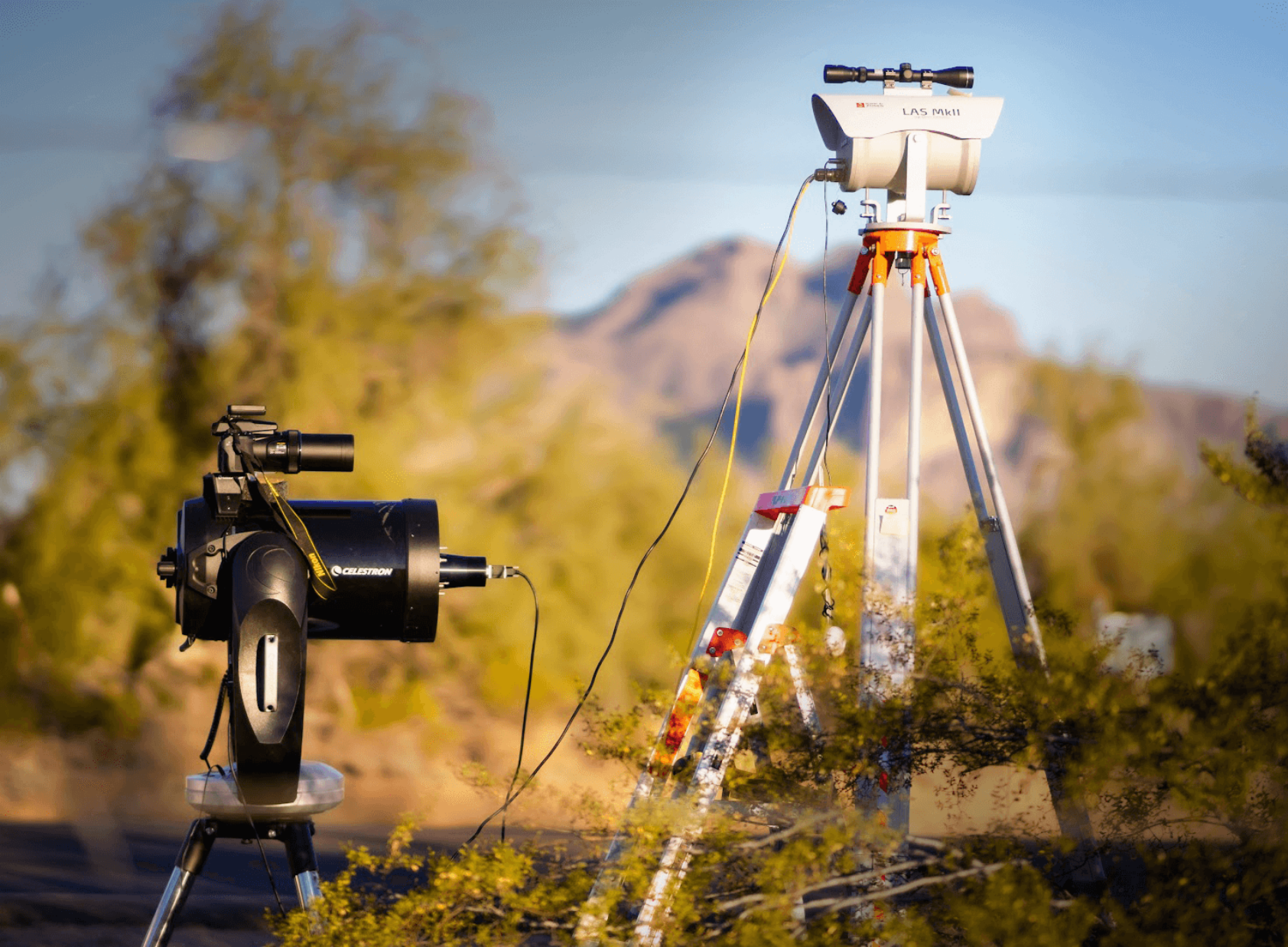}
    \caption{Nikon camera on top of telescope (left side) and scintillometer (right side)}
    \end{subfigure}
    \begin{subfigure}{0.49\columnwidth}
    \centering\includegraphics[width=.99\columnwidth]{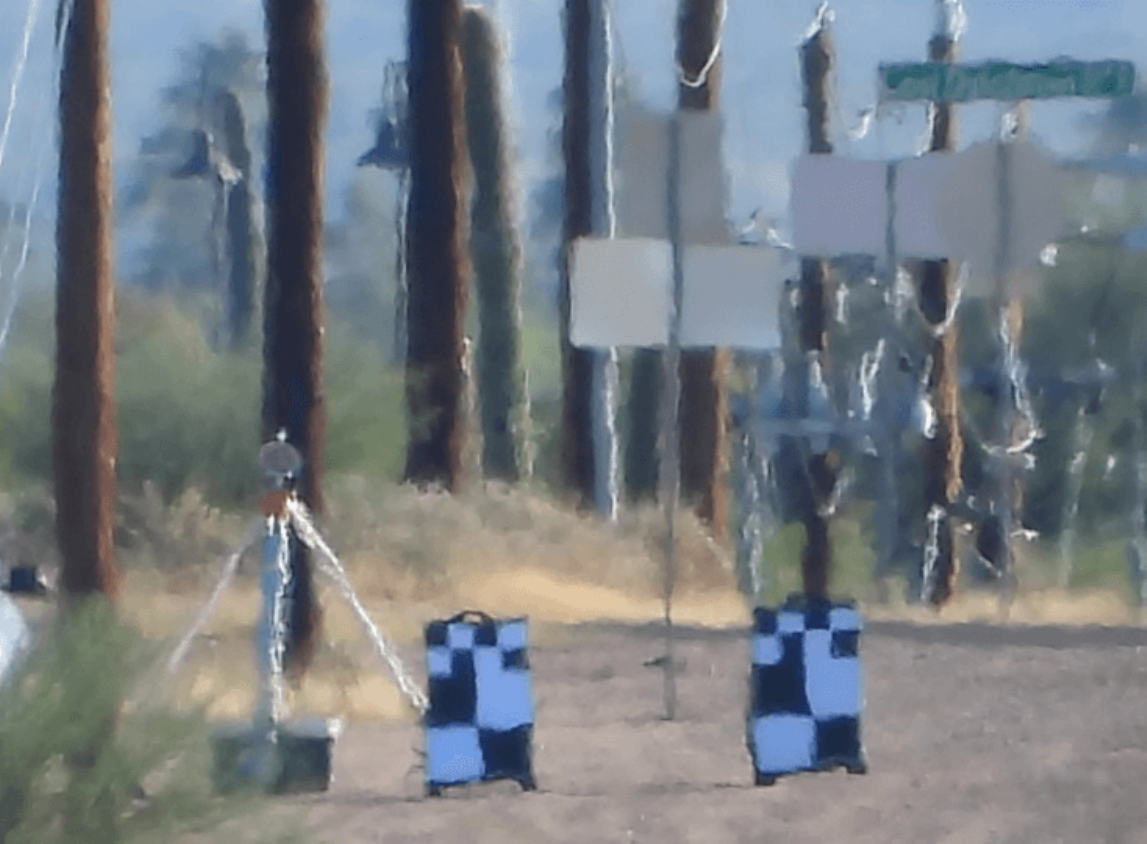}
    \caption{A cropped portion of the original images containing target}
    \end{subfigure}
    
    \caption{(a) The scintillometer receiver is measuring $C_n^2$ values, and images are captured with a Nikon zoom camera and Celestron telescope. (b) The image captured by the camera shows two target boards and the scintillometer. This distortion and bending of lines in these captured images are due to atmospheric turbulence.}
    \label{fig:experiment}
\end{figure}

 To train/test any machine learning models to predict $C_n^2$ from images, we need both long-range images and $C_n^2$ values from the scintillometer from the same location with the same line of sight at the same time. For these experiments, we have collected data of $C_n^2$ using a scintillometer and images using a camera with a super-telephoto zoom range. Also, we collect some weather information required to calibrate the scintillometer for accurate measurement. We conducted two data collections in two different locations in the greater Phoenix, Arizona area.


\textit{Setup:} We have used the following instruments in our experiments: (1) A Kipp \& Zonen LAS MkII Large Aperture Scintillometer; (2) Nikon P1000 with an effective focal length of 3000mm, Resolution of 1920x1080p and frame rate was 120fps; (3) a prototype weather station used on the receiver side to collect essential weather data like temperature, humidity, and wind speed; (4) custom printed target boards were used to display square grids of black or white color, set up on the transmitter site of the experiment. The turbulence effect is visible on those squares as those grids get distorted with high turbulence.


All instruments were set up on either of two sides of the experiment: The transmitter and receiver side of the scintillometer. On the transmitter side, we had a scintillometer transmitter to send the signal to the receiver scintillometer with a large distance between them. Also, the target board was set up on the transmitter side so that cameras from the receiver side can capture the image of this target board and measure the image distortion caused by turbulence. The weather station was set up on the receiver side. When setting up the scintillometer, those are aligned with the viewfinders of the scintillometer and calibrating the power.

\textit{Site Location:} Our first data collection was performed on July 2021 in the Usery Mountain Regional Park, Arizona with a 724m distance for 3 hours with a single target board. The second set of data was collected at Apache Junction, Arizona on October 28, 2021 for 2 hours and 45 minutes during the evening. Two target board was used to capture the data from a distance of 1.45 km with a similar setup as Figure~\ref{fig:experiment}.



\subsection{Data Capture and Preprocessing}
We have captured $C_n^2$ values from receiver scintillometer devices connected to computer. Also, images were captured by Nikon P100 over time by connecting the camera to the same computer. Unlike conventional approaches where the scintillometer and imaging devices are set up at different places with different distances, we set up all the instruments at two points, either the transmitter side or receiver side, which provide more accurate correspondence.  

\textit{Data Capture:} 
We collect scintillometer values in form of $C_n^2$ (minimum value, maximum value, normalized value, and standard deviation) for every 1-minute interval. We have captured 120 images each minute at 120fps with 1-minute intervals. \revise{The camera was set to auto exposure to avoid overexposure and underexposure while collecting data for a long period of time. Gradient methods require image sequences with relatively short exposure~\cite{tofsted2014extended}. In our data, most exposures were less than 8.33ms, while the maximum exposure taken was 16.66ms (taken during the evenings with low illumination), which follows similar exposure times reported by other gradient approaches~\cite{zamek2006turbulence}.} All images contain the metadata to the timestamp so that images can be synchronized with scintillometer values.

\textit{Motion Stabilization:} 
Images are often affected by camera shake and miss-alignment in long-range imaging. \revise{In Section~\ref{sec:sim}, we show the effects of motion on simulated data to illustrate the need for compensating for this motion.} To solve this issue we have approached a two-step procedure. Initially, we manually selected a fixed position of the target board of all images over time and aligned all the images together based on that point. The issue with drifting and large motions was resolved with this procedure. However, some vibration and high-frequency motion were still visible after that correction. We solved that by selecting image features between frames, finding features correspondence between images and transforming images to align all of them. The resultant image sequences were consistent without observable vibration.

\subsection{Method Implementation}
We discuss three main methods we implement for $C_n^2$ estimation on our collected data: (1) the gradient method of Zamek et al.~\cite{zamek2006turbulence}, (2) our own CNN architecture for directly regressing turbulence strength, and (3) our hybrid physics-based CNN coupled with the differentiable gradient method.

\subsubsection{Gradient-based $C_n^2$ Estimation}
Within limited research performed to compute $C_n^2$ from recorded image sequences, the gradient approach~\cite{zamek2006turbulence, zamek2006TurSuper, porat2011optical, gladysz2013estimation} is considered to be the most popular approach. We implement the gradient-based approach based on Zamek et al.~\cite{zamek2006turbulence} in the following manner. 

Let X be a sequence of images captured from an RGB sensor, $PFOV$ as pixel field of view in radians ($5.95\times10^{-6}$ in our camera), D as the diameter of the lens aperture ($0.06$ meter), L as the distance from the camera to target ($724 \sim 1450$ meter depending on the scene) and P as the turbulence constant parameter (usually 1.1 or 2.9). Then $C_n^2$ estimation from the image gradient can be formulated as: 

\begin{equation}
    C_n^2 = \frac{PFOV^{2} \times D^{\frac{1}{3}}}{L\times P} \times \frac{\sigma (X)^{2}}{Grad(X)}. \label{eq:1}
\end{equation}

\subsection{Deep Learning for $C_n^2$ estimation}

\begin{figure}[h!]
    \centering
    \includegraphics[width=.99\columnwidth]{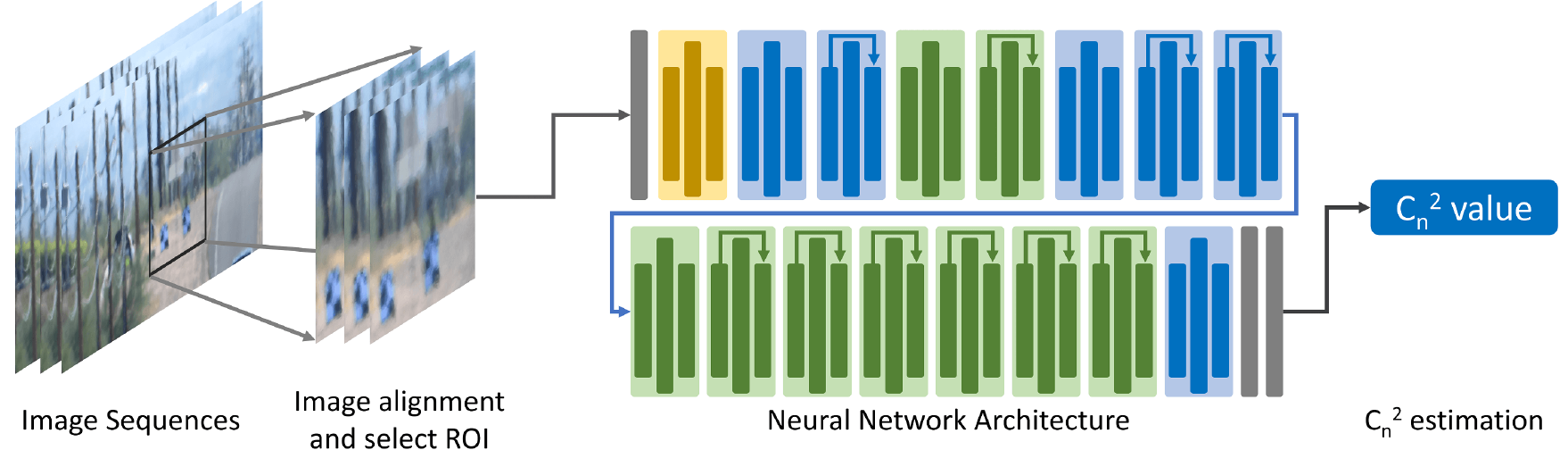}
    \caption{Overall \revise{deep learning architecture} for $C_n^2$ estimation. Initially, all images are aligned together to remove high-frequency motion, and the region of interest was cropped out. Using an EfficientNetV2 inspired architecture, we estimate the $C_n^2$ value in supervised learning. \revise{Here, the gray blocks represent a Conv 3x3 layer with Batch normalization and Swish activation, the yellow block represents a Fused-MBconv1 architecture~\cite{tan2021efficientnetv2} with 3x3 kernel, the blue block represents the Fused-MBconv6 architecture with 3x3 kernel, and the green block represents the Fused-MBconv6 architecture with 5x5 kernel. The Fused-MBconv architecture itself consists of a Conv3x3 layer with a Squeeze and Excitation(SE) Layer followed by a Conv1x1 layer with skip connections.
}}
    \label{fig:OverallPipelien}
\end{figure}

We have used a neural network architecture as shown in Figure~\ref{fig:OverallPipelien} based on EfficientNetV2~\cite{tan2021efficientnetv2} for $C_n^2$ prediction from images. This model was discovered based on a neural network search to provide fast training with efficiency and accuracy in the field of computer vision. This architecture can easily handle different image sizes. For this experiment, we have selected image patches of 256 $\times$ 256 from the target image. However, we have passed 3 consecutive images in the model by converting them into grayscale images. The output of the neural network is the $C_n^2$ value.


The network architecture uses a specialized block called Fused-MBConv which uses convolutional operation with a 3x3 window, attention block to look at a portion of the image, and convolutional operation of 1x1 that works on a filter map of images. After all the convolutional operations, we get a 7$\times$7$\times$30 size matrix which was passed into a fully connected network to get the prediction value of $C_n^2$. For this experiment, we have divided the data into training and testing sets where the testing set contains 140 batches of images for comparison and training set contains 8680 batches of images. We have added the batch size of \revise{3}, so at a single time, we pass \revise{3} \delate{batches of} images along with corresponding scintillometer values into the network architecture to train the model faster.

\textit{Implementation details:} 
We have 120 images collected each minute for our data. \revise{It's possible to provide a relatively short sequence of images {$3 \sim 10$} images as input to the model.} We have selected \revise{3} \delate{$3 \sim 10$} images as a group to pass into the CNN. So, we have \delate{$12 \sim 40$ } \revise{3} $C_n^2$ values predicted each minute from the CNN compared to a sampling frequency of 1 from scintillometer. To adjust for this mismatch, we compute the median of the CNN predictions for comparison. To speed up the image loading process, we have cropped all images to $256 \times 256$ size output with all metadata and timestamp saved. We trained the model on an NVIDIA Titan XP GPU with a learning rate of $2\times10^{-6}$ for 30 epochs, which took about 2 hours of training. 

\subsection{Physics-based CNN for $C_n^2$ estimation}
Deep learning models are best suited when the scene location, camera, and direction are fixed and $C_n^2$ need to be estimated with the fixed optical setup for multiple days. However, when camera parameter changes such as the focal length, sensor size, and aperture, then deep learning models do not generalize in those cases.

\begin{figure}[h!]
    \centering
    \includegraphics[width=.99\columnwidth]{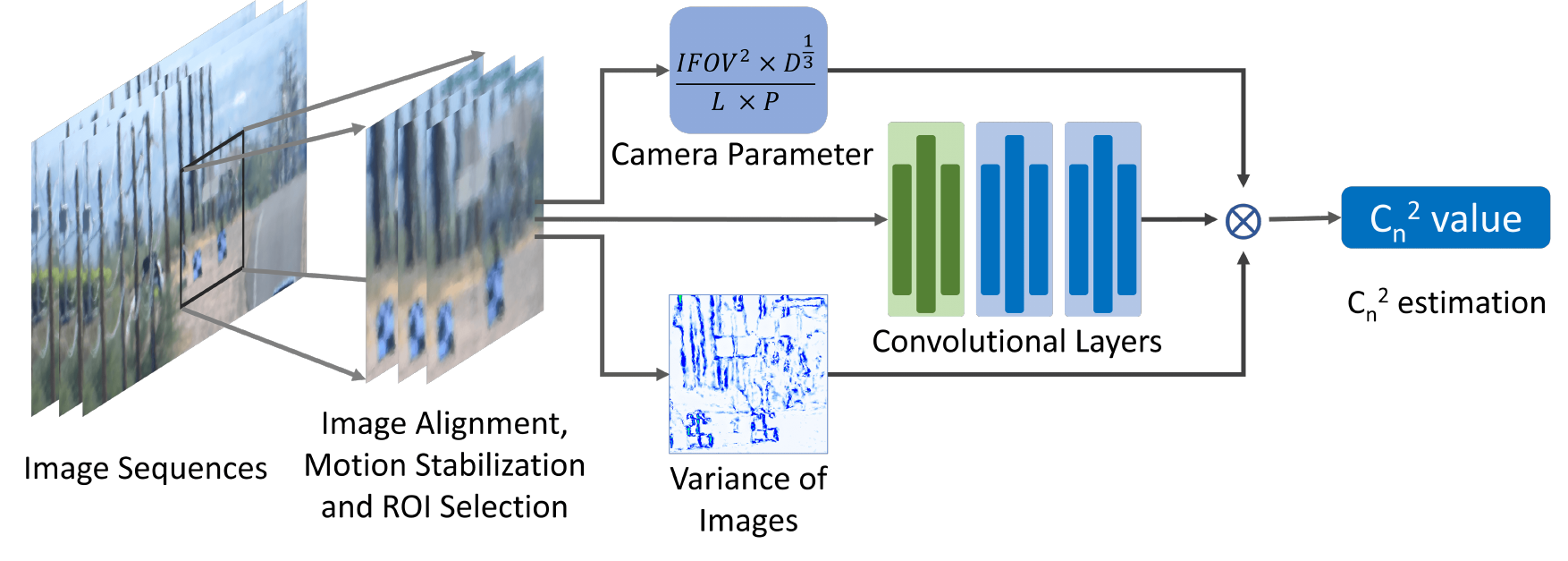}
    \caption{Physics-based deep learning to estimate $C_n^2$ from image sequences with different camera parameters. \revise{The green block is a single convolutional layer with a 5x5 kernel and depth 3, and the blue blocks are convolutional layers with a 5x5 kernel and depth 1. The output of the convolutional layers is then multiplied by the camera parameters and the image variance as specified in Equation~\ref{eq(2)}.}}
    \label{fig:differentiable Gradient Pipeline}
\end{figure}

To address this lack of generalization, we have developed a physics-based deep learning model as shown in Figure~\ref{fig:differentiable Gradient Pipeline} that is inspired by the image gradient algorithm~\cite{zamek2006turbulence} and can implicitly focus on the variance of images over time to calculate the $C_n^2$. We have represented the Equation \eqref{eq:1} as follows:
\begin{equation}
    C_n^2 = M\times \frac{Var( I)}{Conv^{n}( I)} \label{eq(2)}
\end{equation}

Here, Var(I) compute the variance of image sequences, and different convolution operation was applied to the image sequences. Those deep learning convolution operators act like filters able to compute gradients just like different edge detection filters. Here, $M$ is a single number and is represented by:
\begin{equation}
    M = \frac{PFOV^{2} \times D^{\frac{1}{3}}}{L\times P}
\end{equation}

Here, all camera parameters are explicitly provided to the model, enabling robustness. During back propagation, the CNN determines the weight of all the convolutional layers to match the $C_n^2$ values. During our experiment, we found keeping the number of images in the range of 2 to 20 provides the best accuracy.

\section{\revise{Simulations}}
\label{sec:sim}

\revise{While previously we have discussed our real data capture, the difficulty of acquiring large amounts of data enables only limited analysis and experiments that we show in Section~\ref{sec:results}. For this reason, we detail our simulation results using a physics-based turbulent image simulator and then show a comparison of gradient method, deep learning, and physics-based deep learning methods for simulated data.}

\subsection{Physics-based Simulator}
\revise{Conventional ray tracing through the 3D volume is computationally expensive and costly, while simplistic tilt-and-blur simulators do not capture the true statistics of the turbulence medium. An alternative to these approaches is to model turbulence as a series of phase screens that models the refractive bending of light~\cite{simaniso}. In particular, the collapsed phase-over-aperture model proposed~\cite{chimitt2020simulating} models these phase screens locally per pixel. While this method is promising, it still had relatively slower rendering times.}

\revise{Recently, a new physics-based simulator has improved the speed of this phase-over-aperture model~\cite {mao2021accelerating}. This model uses convolutions with learned basis functions combined with a learned phase-to-space transform to accelerate traditional simulators by 300x - 1000x and enables fast simulation of images necessary to generate datasets. We leverage this simulator for our experiments below.}

\subsection{Sensitivity to Motion}
\revise{In our first simulated experiment, we sought to simulate the effects of motion on the three proposed methods. To do so, we simulated a dataset with 24 different scenes for training purposes, and 24 different scenes for testing. We swept $Cn^2$ from 1e-16 to 1e-11 to simulate a full range of turbulence strengths for these images. For each image, we simulated motion with pixel displacement in any 2D direction that ranged from 0 pixels (no motion, ideal case) to 8 pixels (severe motion).} 

\begin{figure}
    \centering
    \includegraphics[width=.99\columnwidth]{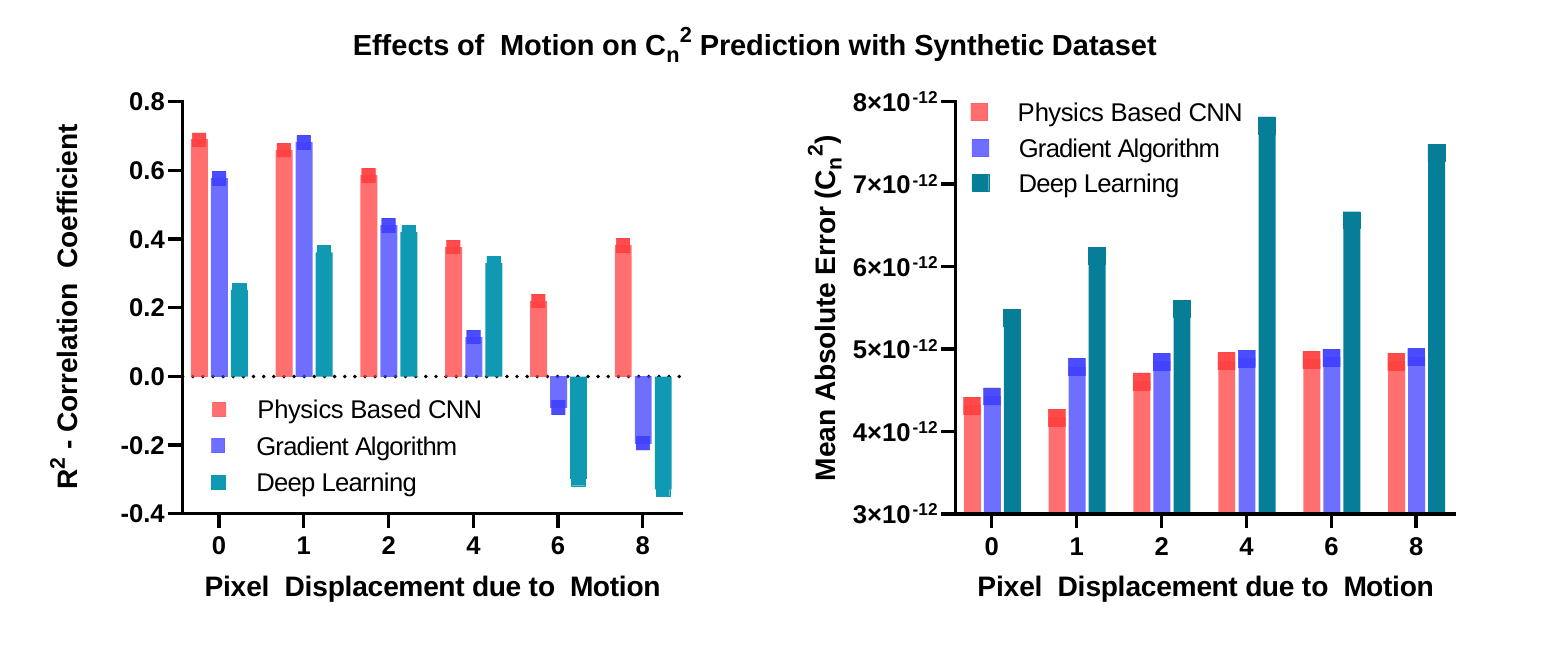}
    \caption{Comparison of the three methods on simulated data with respect to camera motion. We report both $R^2$ and MAE for these methods.}
    \label{fig:Synthetic_Effects_of_Motion}
\end{figure}

\revise{In Figure~\ref{fig:Synthetic_Effects_of_Motion}, we show the results of our experiment in both correlation coefficient ($R^2$) and mean absolute error (MAE). Note that we find the correlation coefficient more informative as this details how well the method followed the trend of the turbulence strength, while MAE can be sometimes difficult to interpret. As we see, the physics-based CNN maintains a high correlation with the ground truth while lower or comparable MAE to the gradient method. Motion affects the gradient and deep learning methods as their performance degrades with pixel displacement.} 

\subsection{ROI Selection}

\revise{In our deep learning implementation, we can only choose ROIs with size $256 \times 256$ due to the fixed convolutional layers at the beginning of our network. For this reason, we chose to study the effect of changing the patch location in an image for the three methods, to see if scene content affects the methods. In this experiment, we simulate 4 images with 4 different patch locations to get a total of 16 training patches at a range of 1E-16 to 1E-11 turbulence strength, and we utilize 4 images with 4 different patch locations from each images with the same turbulence range for testing.} 

\revise{Our results were as follows (averaged over the 16 test patches): Gradient method ($4.38\times 10^{-12}$ MAE, $R^2$ = 0.673), base deep learning ($6.89\times 10^{-12}$ MAE, $R^2$ = 0.382), and physics-based CNN ($4.63\times10^{-12}$ MAE, $R^2$ = 0.772). Note that the physics-based CNN and gradient methods are roughly equivalent in MAE, but the physics-based CNN has a slightly higher $R^2$ which means it correlated higher with the data. The deep learning method again did not perform as well in either metric, again due to a lack of generalization to unseen ROI patches in the test data.}

\subsection{Effect of Aperture}
\begin{figure}
    \centering
    \includegraphics[width=.99\columnwidth]{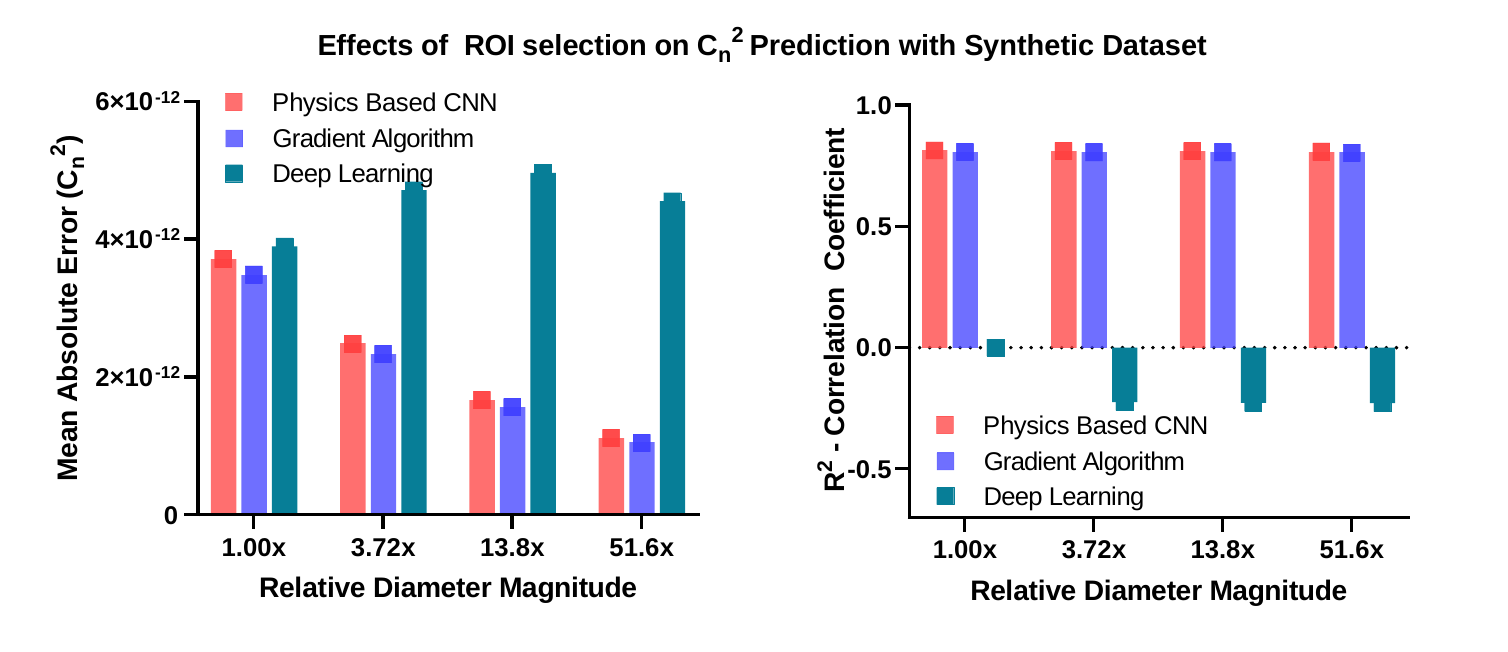}
    \caption{Here the diameter was varied by 1 to 51.6x \revise{in simulation}, to observe the effect of error by different approaches. Here gradient methods show the best accuracy, while the physics based CNN shows very close results, while the \delate{naive} deep learning approach shows large errors representing its difficulty generalizing to different apertures. It also shows that the physics-based deep learning has better correlation while the \delate{naive} deep learning method provides the worst generalization with negative correlation numbers. }
    \label{fig:aperture}
\end{figure}

\revise{We also investigated the effect of capturing data with a different aperture to see how the methods generalize to changes in that optical parameter. With the simulator, we generated a series of $C_n^2$ values from within a range of $1\times10^{-16}$ to $1\times10^{-11}$. In Figure~\ref{fig:aperture} you can see the results of the methods with respect to changing aperture. The physics based CNN shows similar errors as the gradient approach, while the \delate{naive} deep learning approach has difficulty generalizing to the different apertures.}

\section{Experimental Results}
\label{sec:results}
\subsection{Gradient Method Results} 
We evaluated the image gradient based methods and their robustness to different factors: implementation of image derivative, platform correction, different ROI selection, and selection of patch size. This analysis selects optimal parameters to fine-tune the gradient algorithm. It also helped inform the development of the physics-based deep learning model for $C_n^2$ prediction.

\begin{figure}[h!]
    \centering
    \includegraphics[width=.99\columnwidth]{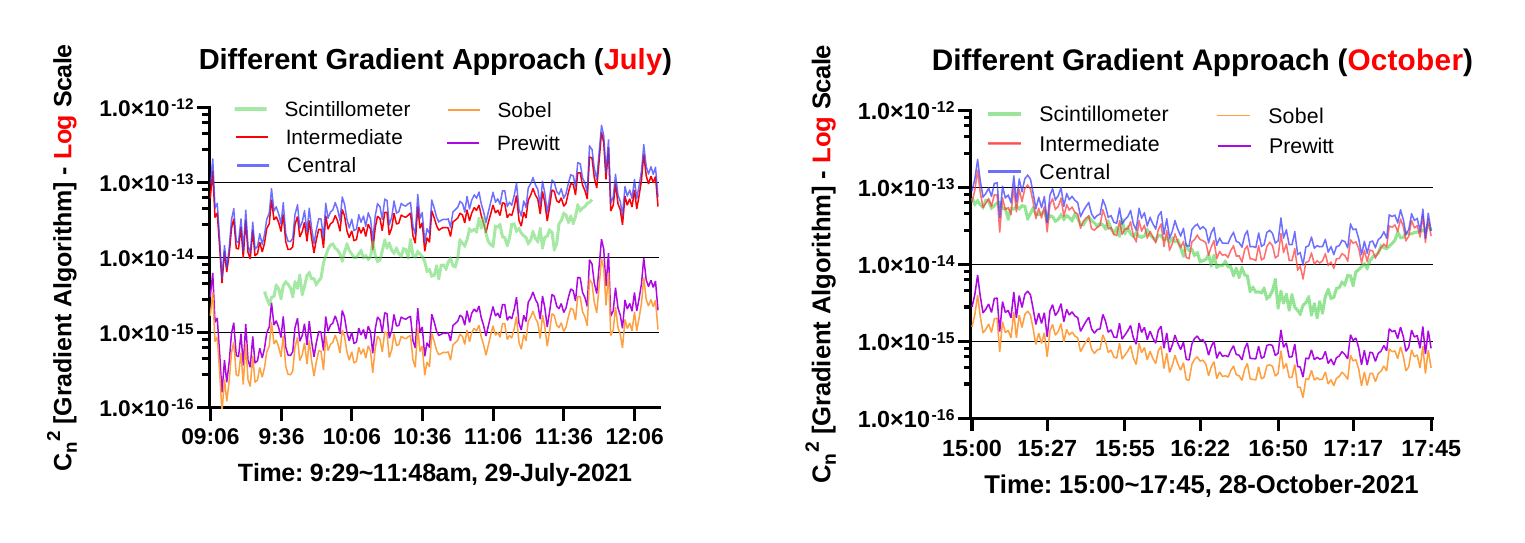}
    \caption{Based on different choices of gradient implementation, existing gradient-based $C_n^2$ estimation approach show high variance on the captured data. The intermediate difference had the highest accuracy for both the July and October dataset 1 (Ref. \cite{saha_dataset}) for estimating the scintillometer measurements.}
    \label{fig:JO Edge Detection.png}
\end{figure}

\paragraph{Choice of Gradient Method} 
Image gradient based methods estimate the variance of the angle of arrival (AOA) of rays coming from the scene to the camera. With higher turbulence, the temporal variance of images also increases. This temporal variance is usually computed by the derivative of pixels in both X and Y directions ~\cite{porat2011optical, o2017strength, zamek2006turbulence, zamek2006TurSuper}. Image derivatives have different practical implementations and are usually computed by kernel/convolution filters applied to the images. The Sobel filter, Prewitt filter, central difference filter, and intermediate difference filter are some of the most popular implementations. However, to the best of our knowledge, we did not find any implementation details of the gradient kernel used in the previous literature~\cite{porat2011optical, o2017strength, zamek2006turbulence, zamek2006TurSuper}. This forms the first comparative point of our analysis. Surprisingly, there is a magnitude of difference in the calculated $C_n^2$ based on just the choice of gradient/derivative implementation, as shown in Figure~\ref{fig:JO Edge Detection.png}. The Sobel and Prewitt-based gradient calculations produce a lower magnitude $C_n^2$ value compared to the scintillometer. However, the intermediate difference and central difference methods were very close to the ground truth, with the intermediate difference giving the best result on the October dataset (a mean absolute error (MAE) of 7.68E-15 as summarized in Table~\ref{table:Paper Ablation Study 3.pdf}, yet a larger difference in the July dataset (3.99E-14 MAE). 

\revise{We also note that the October dataset has a dip in the $C_n^2$ values which the gradient method seemingly does not capture well. There are two potential explanations for this phenomenon. First, the effect is highly visible because of visualization in log scale while in the linear case (not pictured in the paper) the gap is much less. Secondly, the illumination in the scene was changing from day to evening during this transition and images were darker with shadows present. Thus small changes in pixels due to the low turbulence strength may not be easily observed in the image which can account for why the gradient method does not track the dip in the curve.}

\begin{figure}[h!]
    \centering
    \includegraphics[width=.99\columnwidth]{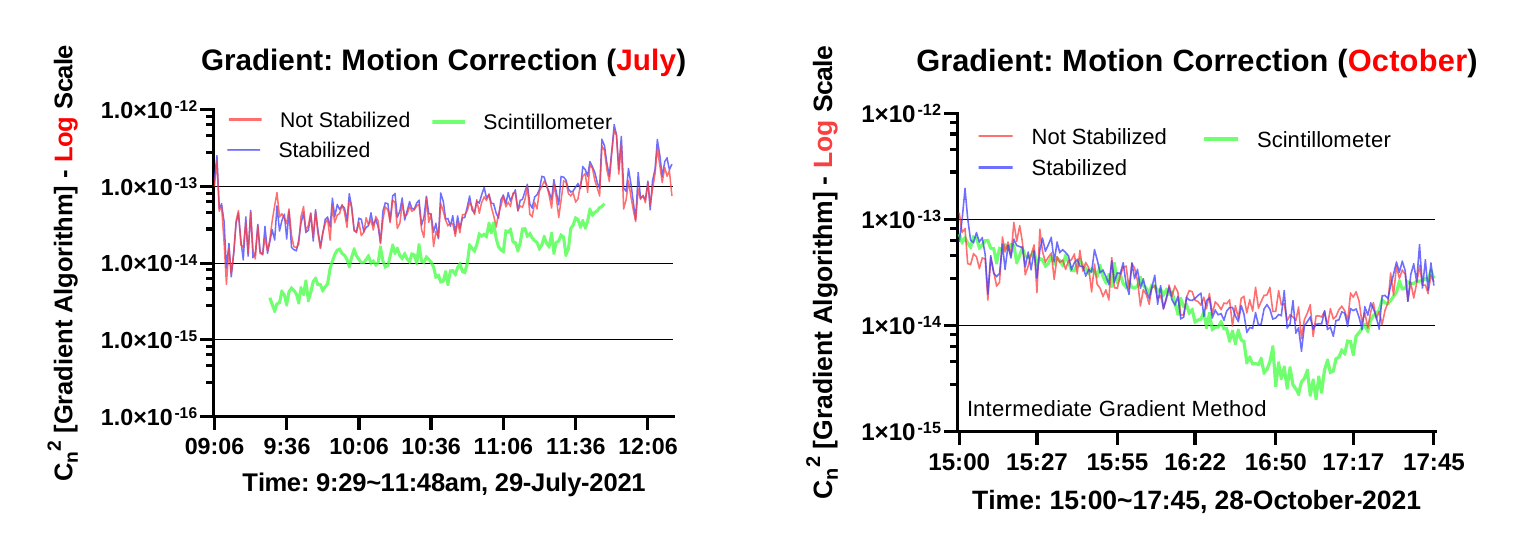}
    \caption{The effect of motion correction for image gradient methods. Overall there is not any noticeable improvement in the July dataset, but we found slightly better alignment of motion corrected $C_n^2$ estimation in the October dataset. }
    \label{fig:JO Platform Correction.png}
\end{figure}

\paragraph{Platform Correction} 
Besides the choice of gradient, other factors can affect the $C_n^2$ computation accuracy. This includes motion induced by camera shake/vibration and movement of objects. This affects image gradient methods which compute the $C_n^2$ by calculating the variance of pixels displacement induced by turbulence. External motion from the camera or object can also displace the images by a large margin in long-distance imaging which adds to the pixel displacement caused by turbulence~\cite{o2017strength}. O'Neill et al. used block matching to iterative find the correspondence block between several frames and partially compensate for platform motion ~\cite{o2017strength}. Zamek et al. used optical flow between two frames~\cite{zamek2006TurSuper} for motion compensation. However, these approaches can hardly deal with severe platform motion and camera drifting. Our two-stage approach to platform correction involves course-scale drifting correction with manual image registration and fine-scale motion correction with image-based feature selection and matching in Figure~\ref{fig:JO Platform Correction.png}. With this approach, we saw only a slight improvement in the October dataset and no visible improvement in the July dataset.

\paragraph{Different ROI Selection}
Selecting different image patches as the region-of-interest (ROI) can also affect the performance of the gradient approach. The main goal is to capture regions with high-intensity fluctuation with high SNR. This is usually done by placing a fixed target with a solid block of black/white pattern ~\cite{porat2011optical, o2017strength}. Another approach is to select a fixed object at a long distance like a building/tower~\cite{zamek2006turbulence, zamek2006TurSuper} with strong edges visible in the image. However, no in-depth comparison has been conducted on the effect of different ROI selections for these algorithms in the prior literature. 

\begin{figure}[h!]
    \centering
    \includegraphics[width=.99\columnwidth]{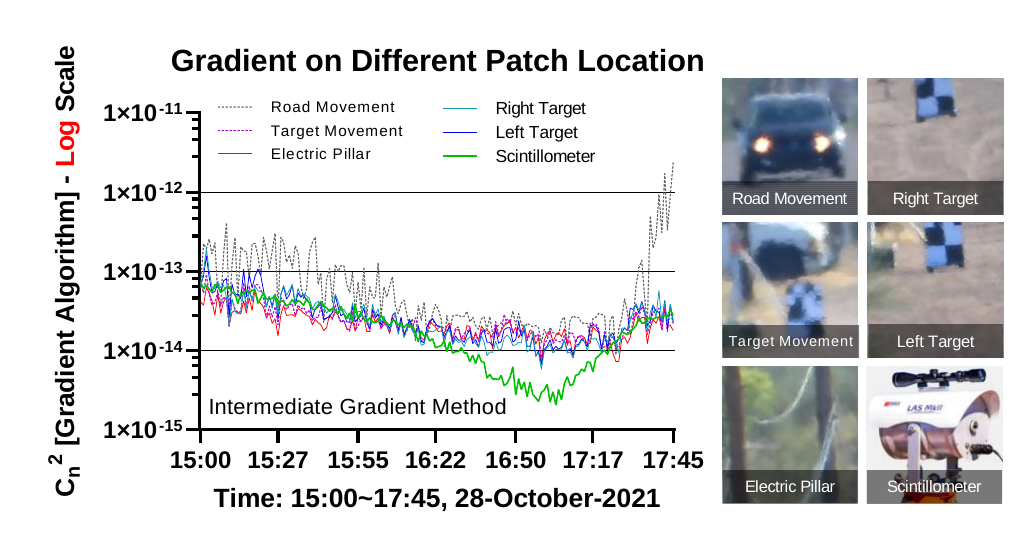}
    \caption{ Comparison of the fluctuation of $C_n^2$ prediction based on different image ROI patches. When there is a movement of objects (car/person) inside images, the $C_n^2$ values fluctuate a lot while fixed patches based on the target board/electric pillar provided much more stable results.}
    \label{fig:DifferentROIwithPicture}
\end{figure}

We have tested the image gradient approach on different ROI (after motion correction). We found that using stable ROIs (target board, or electric pillar) provides much stable $C_n^2$ prediction. ROI with movement (when a car moves on the road or when ROI includes both the target and the road with moving cars) provides unstable prediction and shows high fluctuation. Even though the electric pillar was static, $C_n^2$ estimation based on this feature showed unstable results. The best results we got were from high contrast targets consisting of the black and white pattern in Figure~\ref{fig:DifferentROIwithPicture}. 

\begin{figure}[h!]
    \centering
    \includegraphics[width=.99\columnwidth]{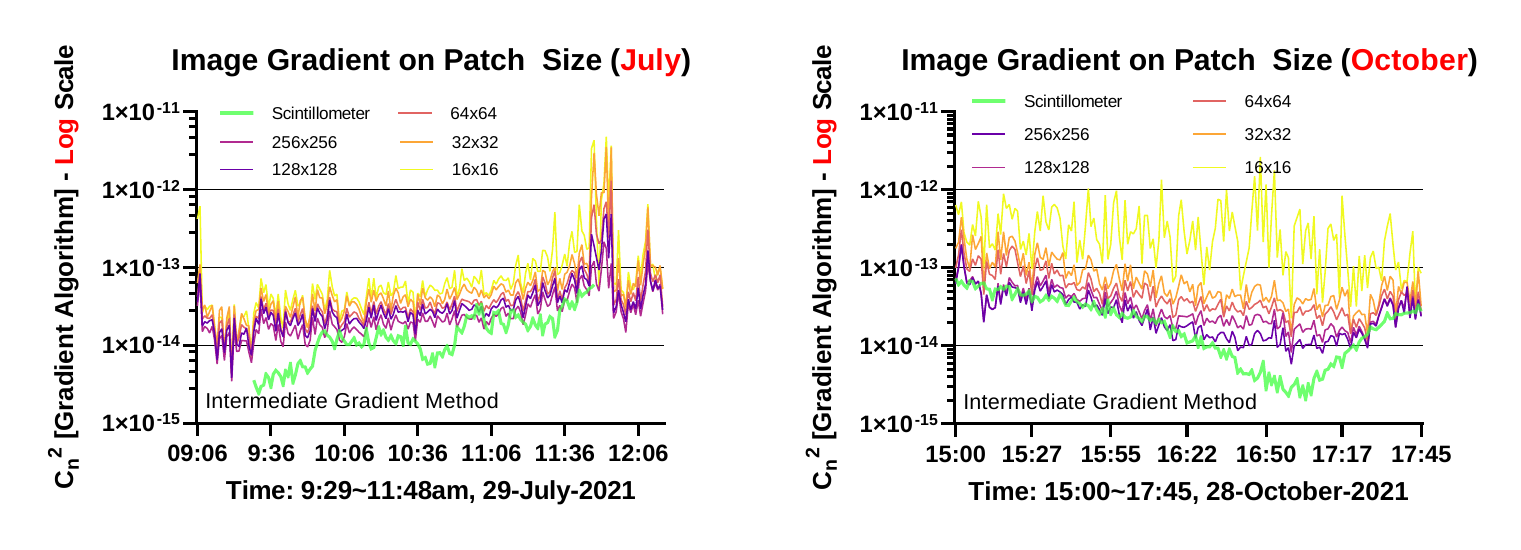}
    \caption{$C_n^2$ prediction from gradient approach based on different patch sizes. Larger patch size seems to provide better results while patch size less than $64\times64$ provided unstable results.}
    \label{fig:JO ROI Patch Size.png}
\end{figure}

\paragraph{Different Patch Size}
ROI size also plays an important role in $C_n^2$ prediction. Prior research typically only considers the region with high contrast that represent strong intensity fluctuations with a sub window~\cite{porat2011optical}. O'Neill et al. used multiple patches of resolution $16\times16$ or $32\times32$ for their study. Other papers selected the whole image for $C_n^2$ calculation ~\cite{zamek2006turbulence, zamek2006TurSuper}. We have experimented with different patch sizes from the fixed target on both the July and October datasets as shown in Figure~\ref{fig:JO ROI Patch Size.png}. We found larger patch size provides closer $C_n^2$ predictions to the scintillometer values. We also observed instability when using smaller patch sizes. Thus, we have used $256\times256$ image resolution as the default size for all further experiments in this paper.

\subsection{Deep Learning Results}
We have also evaluated our base deep learning model based on EfficientNetV2 for the July and October dataset. When testing the model, it's important to represent the performance of the model within the dataset and how well it generalizes to the other dataset. Thus we use three main evaluation methods for testing: 

\begin{itemize}
  \item \textbf{Interpolation Method:} Interpolation measures the network's ability to predict the $C_n^2$ given it has seen training data near the time period of interest. To do so, we divided all data sequentially, and sample sets of images were assigned to training (66\%) and testing (33\%) sequentially. This is the easiest way of testing the neural network since it splits the data randomly into training and testing. This means that the network will likely have seen/trained on a similar set of images from a time period close to the period of interest that is being evaluated. \revise{While this scenario may seem not useful, there can be times where the optical camera is able to be calibrated with a co-located scintillometer (i.e. a fixed optical setup) such that the scintillometer can capture data only periodically to save power while operating the camera continuously to fill in the missing $C_n^2$ values.}

  
  \item \textbf{Extrapolation Method:} Extrapolation measures the ability of the network to predict turbulence strength for extended time periods where it has not been trained on similar data in that time. To implement this, we utilize 6-fold cross-validation by training the network on 5 folds and testing on the sixth one. This helps test the network's ability to generalize to the unseen time fold within a dataset. In other words, the network will not have seen training data within the fold, and thus must generalize its knowledge from other folds to test on this time period. \revise{This scenario can occur in the field when a scintillometer is only used for initial setup and calibration with the camera, and then removed to let the camera operate alone.}

  \item \textbf{Model Transfer: } The final method tests the network's ability to generalize out of the dataset. In this approach, the model is trained with the July dataset and then the performance of the model is tested with October dataset or vice versa. This is the ultimate challenge as the network experiences different images from a different location and time of year in training versus testing. \revise{This is the most useful operating case and the most significant experiment, as it allows the camera and scintillometer to collect data from one location for calibration, and then the camera can be moved to another location without sacrificing performance.}   
\end{itemize}

\subsubsection{Network Performance}
\begin{figure}
    \centering
    \includegraphics[width=.99\columnwidth]{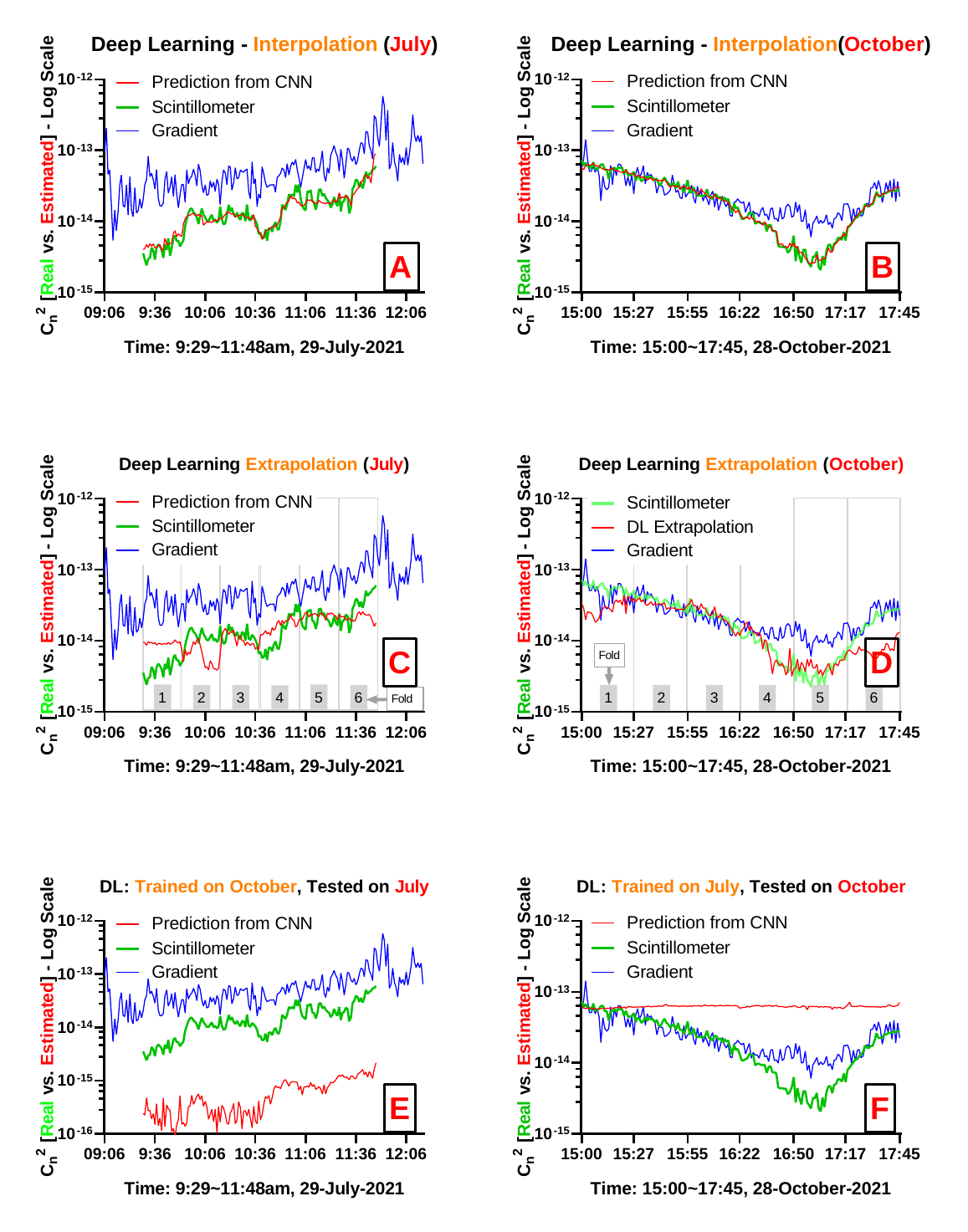}
    \caption{Evaluation of our baseline Deep Learning model. \textbf{A/B [Interpolation]}: The gradient approach was roughly following the same direction as the scintillometer values, but shows a large offset, especially in the case of the July dataset. However, the base Deep learning model can accurately estimate $C_n^2$ in these cases. \textbf{C/D [Extrapolation - K-fold cross validation]}: The results are shown for K-fold cross-validation. As one can see, the deep learning model performs slightly worse on the testing fold as visualized in the figure. \textbf{E/F [Model Transfer]}: The base deep learning model completely fails to generalize when the model was trained on the July dataset and tested on the October dataset because the October dataset does not have the dynamic range of $C_n^2$ that exists on October dataset. When the model is trained on the October dataset and tested on the July dataset, it can slightly estimate the trend of $C_n^2$, however, it still shows a large offset with a large error.
    }
    \label{fig:JO Effnet Full.pdf}
\end{figure}

When training and testing the model by subsampling images between the whole range of the dataset (interpolation), the network performed well compared to the image gradient based approach (from Table~\ref{table:Paper Ablation Study 3.pdf}: the network achieved 2.95E-15 MAE for the July dataset and 1.96E-15 for the October dataset, while the image gradient achieved 3.99E-14 and 7.68E-15 respectively). These results are visualized with respect to the ground truth scintillometer values for both July and October datasets in Figure~\ref{fig:JO Effnet Full.pdf}(A,B). 

\subsubsection{Network Generalization} 
With limited data, it is really hard to generalize a deep learning model. Deep learning models with large parameters can overfit the full dataset. Our interpolation experiments show that our models can achieve high accuracy when the camera position is fixed and the same object is captured for a long time. However, our extrapolation and model transfer tests are designed to investigate how well our model generalizes.



In the case of extrapolation using k-fold cross validation, we found that deep learning struggled to predict the initial and last folds of the data. In Table~\ref{table:Paper Ablation Study 3.pdf}, we see the deep learning had extrapolation errors of 6.6E-15 MAE for the July data and 8.16E-15 MAE for the October data, which are higher than the interpolation counterparts. This can be due to the fact that the deep learning model was not trained with a wide range of $C_n^2$ values to predict such low or high values ~\cite{balestriero2021learning}. However, for other folds, the deep learning model still showed superior results compared to the gradient approach as visualized in Figure~\ref{fig:JO Effnet Full.pdf}(C,D).


The last experiment to test the generalization of deep learning was based on training and testing on different datasets. In Figure~\ref{fig:JO Effnet Full.pdf}(E,F), we see that the model does not generalize well outside of its dataset. In Table~\ref{table:Paper Ablation Study 3.pdf}, the Deep learning model transfer results are the highest with 1.547E-14 MAE for the July dataset and 3.844E-14 for the October dataset.

\subsection{Physics-based CNN with Differentiable Gradient Method}

\begin{figure}[h!]
    \centering
    \includegraphics[width=.99\columnwidth]{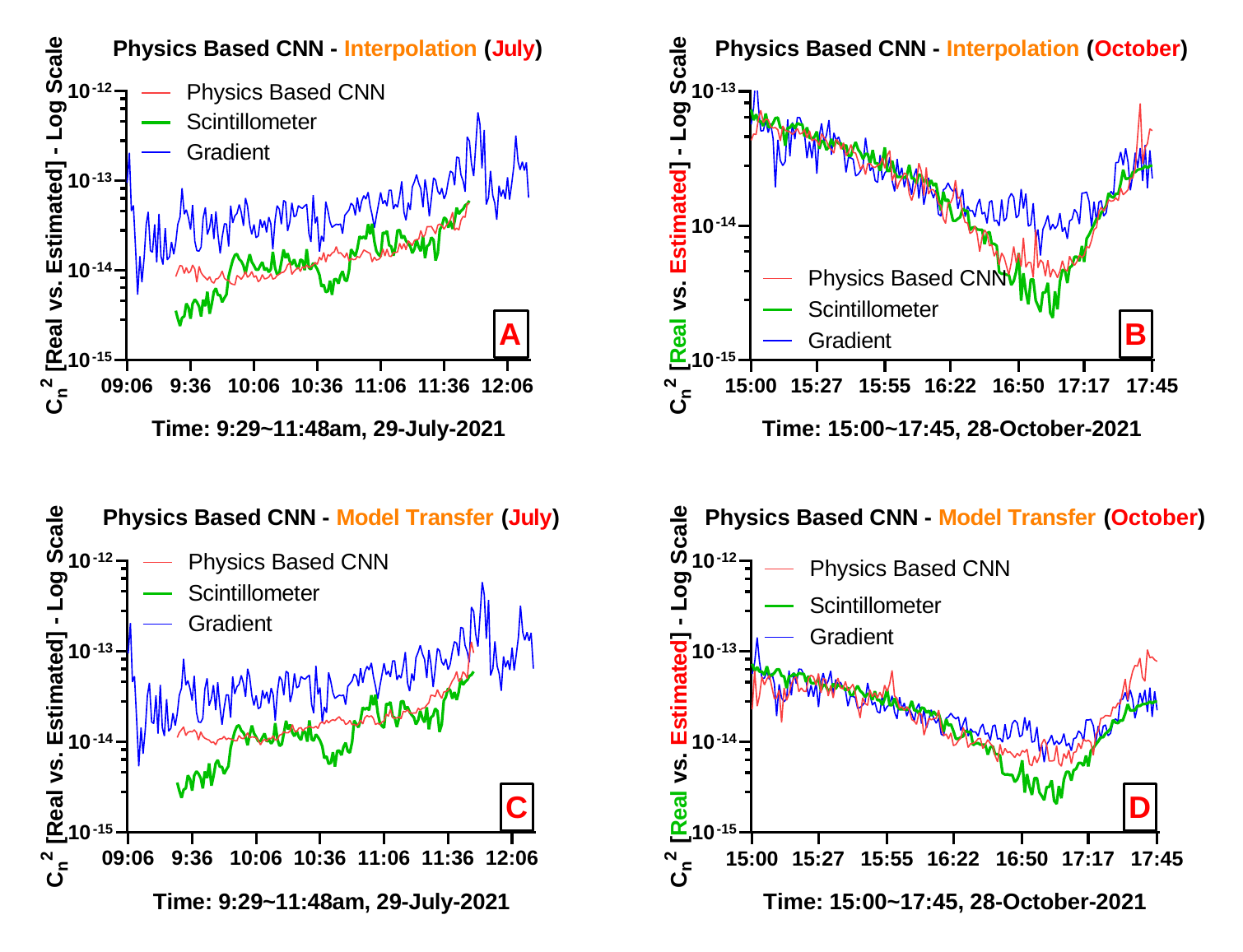}
    \caption{\textbf{Evaluation of Physics-based CNN.} Top(Interpolation): The estimated $C_n^2$ values using physics-based CNN shows that it can predict the trend of $C_n^2$ much better than the image gradient approach. Bottom(Model Transfer): Here the physics-based model was trained on October while tested on July(left), and again the model was trained on July from scratch and tested on the October dataset(right). Note the method gets better performance than the gradient methods while not suffering overfitting that was present with the base deep learning model.}
    \label{fig:JO DiffGrad.pdf}
\end{figure}

Based on our observations about generalization, we designed our physics-based CNN to help improve generalization performance across datasets. In Figure~\ref{fig:JO DiffGrad.pdf}(A,B), we see that in the case of interpolation, the performance of the physics-based model did not overfit the data like the base deep learning model.

We have tested the generalization of this physics-based CNN via model transfer as well. When the model was trained on October data and tested on the July dataset, the generalization is much better compared to the image gradient approach in Figure~\ref{fig:JO DiffGrad.pdf}(C,D), with a MAE of 6.42E-15 in Table~\ref{table:Paper Ablation Study 3.pdf}. Also, when the model is trained on the July dataset, the model was still performing well in the case of the October dataset (MAE of 9.72E-15 in Table~\ref{table:Paper Ablation Study 3.pdf}), with a much lower dip (when scintillometer $C_n^2$ values were lower) and rising curve when the $C_n^2$ values were higher. In comparison, the image gradient methods were relatively flat, and the values were not going down in the case of low $C_n^2$.

\subsection{Quantitative Comparison}

\begin{table}[h!]
    \centering
    \includegraphics[width=.90\columnwidth]{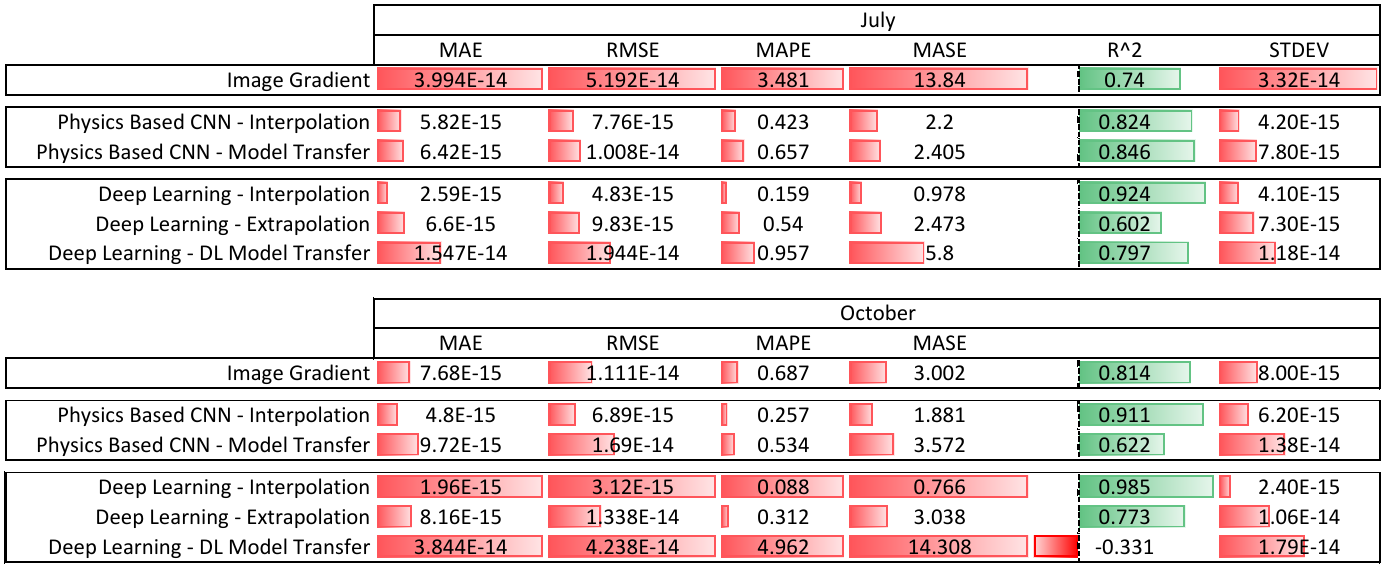}
    \caption{Quantitative evaluation for the image gradient, deep learning, and physics-based CNN methods. However, values are better for the metrics. It can be seen that the image gradient approach has the most difficulty in $C_n^2$ prediction in the metrics specifically in the case of the July dataset. The Physics-based CNN provides the most consistent result for both the July dataset and October dataset considering both interpolation and model transfer results.}
    \label{table:Paper Ablation Study 3.pdf}
\end{table}

An overall quantitive comparison between the image gradient approach, deep learning method, and the hybrid physics-based CNN is summarized in Table~\ref{table:Paper Ablation Study 3.pdf}. Metrics include mean absolute error (MAE), root mean square error (RMSE), mean absolute percentage error(MAPE) and mean absolute scaled error(MASE)~\cite{hyndman2006another}, \revise{correlation coefficient ($R^2$, refereed as linearity in other literature} and STDEV (standard deviation of the error).
The comparison of those approaches based on both the July and October dataset 1 (Ref. \cite{saha_dataset}) is represented in table \ref{table:Paper Ablation Study 3.pdf}. Deep learning interpolation shows the best possible results. However, in the case of model transfer and extrapolation, the neural network performance degrades a lot. The image gradient method seems to provide the least accuracy in the case of $C_n^2$ calculation. Overall, the physics-based CNN provides the most stable results \revise{in case of July dataset, while show similar accuracy in case of October dataset. We anticipate the July dataset doesn't have large range of $C_n^2$ to train the physics based deep learning model to make it robust enough.} \delate{both in the case of July and October detests with interpolation.} Also, during cross dataset generalization, the performance of physics-based deep learning overcomes base deep learning model by a large margin.


\section{\revise{Discussion}\delate{Summary}}
We have conducted a comparative study between classical image gradient methods and deep learning for $C_n^2$ estimation. \revise{As part of our contributions, we capture and analyze two image datasets: dataset 1 (Ref. \cite{saha_dataset}) with 30K+ images each with corresponding ground truth scintillometer values. We have made this dataset  1 (Ref. \cite{saha_dataset}) and code 1 (Ref. \cite{saha}) available to the public, which is the first open dataset of images and scintillometer measurements to the best of our knowledge.}

\revise{In our simulated data, we found the classical gradient method performs slightly better than both of the deep learning methods in an ideal, noiseless setting, but starts to degrade as camera motion/shake affects the imagery. The neural networks both have slight robustness to misalignments in the pixels due to displacement, but even simulated data show that the deep learning model suffers from overfitting and does not generalize during model transfer (i.e. trained on one set of data, tested on another) well. The physics-based CNN can perform model transfer better and generalize to the unseen simulated data. In addition, the choice of ROI location, as well as optical parameters such as aperture, do not affect the physics-based CNN nearly as much as the deep learning base model.}

\revise{In real experimental data,} we found the classical gradient methods show high variation in performance based on different factors such as choice of gradient \revise{implementation}, \delrev{ROI,} patch size, and motion correction. \revise{On the other hand, the base deep learning model is highly accurate for interpolation experiments (i.e. where the model has seen $C_n^2$ values at the given location spanning an entire range) but does not generalize well in the extrapolation or model transfer experiments. The hybrid physics-based CNN with a differentiable gradient approach splits the difference between these methods and can provide accurate results with good generalizability on the dataset.}\delrev{We propose two deep learning solutions to achieve better performance for this task. The base deep learning model is proposed for continuous $C_n^2$ measurement from a fixed target by a video recorder (where the model needs to be trained once). This method is highly accurate but overfits easily to training data. To combat this, we design a physics-based CNN with a differentiable gradient approach to predict the turbulence strength in different settings with strong generalization ability. We show that, even after the image gradient methods are fine-tuned with proper parameter selection, our proposed physics-based CNN can overcome the limitations of image gradient methods by a large margin on the real dataset. }

\textbf{Limitations:}
One of the main limitations of the base deep learning model is poor generalization capability. However, the better interpolation results represent that, when the target is fixed and the model is trained with multiple-day data (containing a high $C_n^2$ dynamic range) from the same location in a constrained environment, the model will be able to predict $C_n^2$ values for future days with better accuracy. In addition, when we tried to train the model with nighttime images, we found there is not enough light to capture details from the images, resulting in low SNR. This prevented the model from working adequately. 

Finally, even though we treat scintillometer data as ground truth in this study, scintillometer data contains a large range of deviations as shown by previous studies. \revise{For instance, the beam-path for the scintillometer is approximately half a meter above the path of the imaging system, which can cause the $C_n^2$ to vary since it is a function of height. We did not measure this height difference along the beam (as the terrain changes) or compensate for it, and thus our ground truth measurements are affected by this.} Thus getting to higher accuracy systems requires dedicated active laser and optical systems to be precisely calibrated.

\textbf{Future Work:}
This work is based on estimating a single turbulence strength number ($C_n^2$) from a set of images. In the future, we plan to estimate spatially-varying $C_n^2$ from each pixel of a complex scene (with depth and temperature variation). This would be useful to help improve turbulence estimation for physical modeling of the environment, as well as other vision analysis tasks for turbulent imagery.

\textbf{Funding:} U.S. Army Combat Capabilities Development Command (W91CRB21C0028); National Science Foundation (IIS-1909192). 

\textbf{Acknowledgments:} This work was supported by the Army SBIR Phase II Contract No. W91CRB21C0028, and NSF IIS-1909192. The authors wish to thank Cameron Whyte for helpful discussions at the start of this project. 

\textbf{Disclosures:} The authors declare no conflicts of interest. 

\textbf{Data Availability:} Data underlying the results presented in this paper are available in dataset 1 (Ref. \cite{saha_dataset}).












\bibliography{sample}
\end{document}